# Deep Learning Methods for Real-time Detection and Analysis of Wagner Ulcer Classification System

Aifu Han
Fujian Institute of Research on the Structure of Matter Chinese Academy of Sciences
Fuzhou, China
School of Data Science and Technology
North University of China
Taiyuan, China
hanaif@mail2.sysu.edu.cn

Yongze Zhang
Department of Endocrinology
the First Affiliated Hospital of Fujian Medical University
Fuzhou, China
Diabetes Research Institute of Fujian Province
Fuzhou, China
274998375@qq.com

Ajuan Li
Fujian Institute of Research on the Structure of Matter Chinese Academy of Sciences
Fuzhou, China
School of Data Science and Technology
North University of China
Taiyuan, China
572731498@qq.com

Changjin Li
School of Software and Microelectronics
Peking University
Beijing, China
lcj@pku.edu.cn

Fengying Zhao
Department of Endocrinology
the First Affiliated Hospital of Fujian Medical University
Fuzhou, China
Diabetes Research Institute of Fujian Province
Fuzhou, China
zhaofyc@163.com

Qiujie Dong
Fujian Institute of Research on the Structure of Matter Chinese Academy of Sciences
Fuzhou, China
School of Data Science and Technology
North University of China
Taiyuan, China
qiujie_dong@mail.sdu.edu.cn

Yanting Liu
Fujian Institute of Research on the Structure of Matter Chinese Academy of Sciences
Fuzhou, China
liuyt@fjirsm.ac.cn

Ximei Shen
Department of Endocrinology
the First Affiliated Hospital of Fujian Medical University
Fuzhou, China
Diabetes Research Institute of Fujian Province
Fuzhou, China
641498885@qq.com

Sunjie Yan*
Department of Endocrinology
the First Affiliated Hospital of Fujian Medical University
Fuzhou, China
Diabetes Research Institute of Fujian Province
Fuzhou, China
fjyansunjie@163.com

Shengzong Zhou*
Fujian Institute of Research on the Structure of Matter Chinese Academy of Sciences
Fuzhou, China
zhousz@fjirsm.ac.cn

***Abstract*—At present, the ubiquity method to diagnose the severity of diabetic feet (DF) depends on professional podiatrists. However, in most cases, professional podiatrists have a heavy workload, especially in underdeveloped and developing countries and regions, and there are often insufficient podiatrists to meet the rapidly growing treatment needs of DF patients. It is necessary to develop a medical system that assists in diagnosing DF in order to reduce part of the workload for podiatrists and to provide timely relevant information to patients with DF. In this paper, we have developed a system that can classify and locate Wagner ulcers of diabetic foot in real-time. First, we proposed a dataset of 2688 diabetic feet with annotations. Then, in order to enable the system to detect diabetic foot ulcers in real time and accurately, this paper is based on the YOLOv3 algorithm coupled with image fusion, label smoothing, and variant learning rate mode technologies to improve the robustness and predictive accuracy of the original algorithm. Finally, the refinements on YOLOv3 was used as the optimal algorithm in this paper to deploy into Android smartphone to predict the classes and localization of the diabetic foot with real-time. The experimental results validate that the improved YOLOv3 algorithm achieves a mAP of 91.95%, and meets the needs of real-time detection and analysis of diabetic foot Wagner Ulcer on mobile devices, such as smart phones. This work has the potential to lead to a paradigm shift for clinical treatment** **of the DF in the future, to provide an effective healthcare solution for DF tissue analysis and healing status.**

# I. INTRODUCTION

Diabetes mellitus (DM) has gradually become an epidemic, the patients with DM worldwide to reach 366 million by 2030 [1]. Diabetic Foot Ulcer (DFU) is one of the most serious and most expensive chronic complications of DM and it may even lead to amputation or death if it is not treated effectively [2]. It is a survey that amputations due to diabetes accounted for 27.3% of all amputations and 56.5% for non-traumatic amputations in 2010 [3-4].

Clinically, the detection and analysis of DFUs is vital in the treatment of diabetic foot [5]. With the rapid popularity of smartphones, more and more researchers focused on developing efficient and inexpensive telemedicine systems to serve for DF patients, such as disease evaluation and teletherapy. Wang et al. adopted an automated ulcer assessment system to measure the ulcer area [6]. The approach provides the patients' Podiatrists the opportunity for asynchronous evaluation. Goyal et al. proposed DFUNet, a novel deep learning framework [7], to distinguish between normal (healthy) skin and abnormal (DFU) skin. This work not only evaluated the performance on the DFU dataset, but also on the facial skin dataset, and report exciting results on both datasets, which present the robustness of the DFUNet performance. Once the patient is diagnosed with a diabetic foot ulcer, a clinical assessment should be performed. Currently, widely accepted grading methods are Wagner grades and Texas grades [8]. The Wagner classification, based on the severity of the ulcer, is composed of six grades. The larger the value, the more serious the degree of the DF. Compared with the Wagner classification, the Texas classification mainly evaluates the DFU and gangrene based on the degree of the lesion and the cause of the disease, which better reflects the wound infection and ischemia [9]. The Wagner ulcers of DFU are summarized as follows: grade 0 (intact skin), grade 1 (superficial ulcer), grade 2 (deep ulcer to bone, tendon, deep fascia or joint capsule), grade 3 (deep ulcer with abscess, osteomyelitis, or osteitis), grade 4 (forefoot gangrene), and grade 5 (whole-foot gangrene) [10]. In most cases, diabetic foot wounds are classified as the presence of open wounds or the absence of open wounds. Base on the clinically diagnosed Wagner grades, for open wounds, bone marrow stroma is assessed using X-rays, magnetic resonance imaging (MRI), or bone scans in order to determine the presence of osteomyelitis. Signs of no open wound or only infection can be distinguished by methods such as color Doppler ultrasound or the Ankle Brachial Index (ABI).

At different Wagner grades stage, the DF skin appearance shows great differences in color and texture. Computer vision algorithms based on the feature of color and texture have been developed to distinguish DF [7]. There is no doubt that the application of computer vision algorithms based on deep learning in medical images is an emerging field, especially in medical image detection and segmentation. Computer vision algorithms based on traditional machine learning mainly rely on feature detection, image preprocessing, feature extraction, feature screening, and inference prediction and recognition to identify DFU or wounds in general [11]. These traditional machine learning algorithms use the difference in color and texture descriptors on the surface of the DFU / wound, and then binary classification (i.e. normal skin patches and ulcer skin patches) is performed using a support vector machine, neural network, random forest or Bayesian classifier [12-14]. In recent research, many researchers have used deep learning-based methods to locate and detect DFUs. Goyal et al employed a real-time detection and localization method for diabetic foot ulcers with 1775 images. [15]. The experiments reported promising results, Faster R-CNN coupled with Inceptionv2 model reached an mAP of 91.8%. The authors used prototype android application to detect foot ulcers on mobile phones. However, their models can only detect whether the foot has ulcers. As far as we know, DFU research based on object detection or image segmentation tasks, whether applying traditional methods or deep learning methods, is limited to binary classification (i.e. normal and abnormal skin pieces) [13, 15]. These methods do not provide a good assessment of the healing state of DFU. For DFU patients who are not willing to be hospitalized, such automatic examination methods can be used for self-management. Wang et al. stated in their paper that the use of digital photography allows patients to observe difficult-to-see wounds [16], which can improve patient adherence to treatment.

According to previous studies, the present wound image-based prediction model focus on the following four tasks: 1) The measurement of wound area and boundary; 2) Wound healing rate assessment; 3) Wound image capture, and 4) Wound localization and classification [17]. This paper focuses on task 4. In this paper, we aimed to develop a system that performs real-time mobile detection, localization and classification according the Wagner grade as a means of assisted screening to alleviate part of the workload for doctors. In addition, the system can also be used as a home telemedicine system to help DF patients perform self-management. This paper's main contributions are as follows:

1) We present the DF dataset, which consists of 2,688 images with the annotated ground truth.

2) This paper uses a deep learning-based object detection algorithm to achieve a multi-class classification of Wagner grades of DF for the first time. To date, the research on DF, whether it is classification, object detection or image segmentation tasks, leads to a binary classification, i.e. normal skin (healthy skin) patches and abnormal skin (DFU) patches. Different Wagner grades require clinically different treatments and wound management strategies. Accurate identification of Wagner grades of DF provides a more accurate assessment of DFU wound tissue analysis and healing status, compared to binary classification.

3) To the best of our knowledge, we are the first to try to utilize stack tricks on medical datasets, and the experimental results achieved are promising – using refinements on YOLOv3 models, the mAP reaches 91.95%. Under smartphone, the speed of inferring from a single picture meets the real-time requirements.

## II. MATERIALS AND METHODS

This section elaborates on the collection of DF data, preprocessing, expert labeling, stack methods used in the model training process, and performance measures of experiments.

### A. DF Dataset

In this paper, all DF dataset were collected from the First Affiliated Hospital of Fujian Medical University (1st. of FMU). All subjects gave their informed consent for inclusion before they participated in the study. The study was approved by the Ethics Committee of FMU Ethics Committee, Approval No. MRCTA, ECFAH of FMU [2017] 131. The DF dataset were obtained by the foot care staff of the hospital's endocrinology department using a dedicated camera at a dedicated podiatric studio. Together with clinicians, we selected 2,535 DF dataset from the hospital image archive, all of which are based on Asian skin color. In order to meet the requirements of rigorous medical studies, different cameras should not be used for capturing DFU images in the healthcare setting, however, in order to ensure our system's robustness, we captured heterogeneous datasets [7]. Ultimately, the model also needs to test and run on a smartphone. Based on the above considerations, in the past year, we collected an additional 152 DFU images in 1st. FMU podiatric studio using an iphone7 smartphone. During the photographic process, the mobile phone was held 30-40cm away from the foot, using a sufficient external light source. In summary, the DF dataset obtained had a total of 2,688 images, including six classes, i.e. grade0, grade1, grade2, grade3, grade4 and grade5, each corresponding to the respective Wagner grade. The distribution of each grade in the DF dataset is shown in Fig. 1.

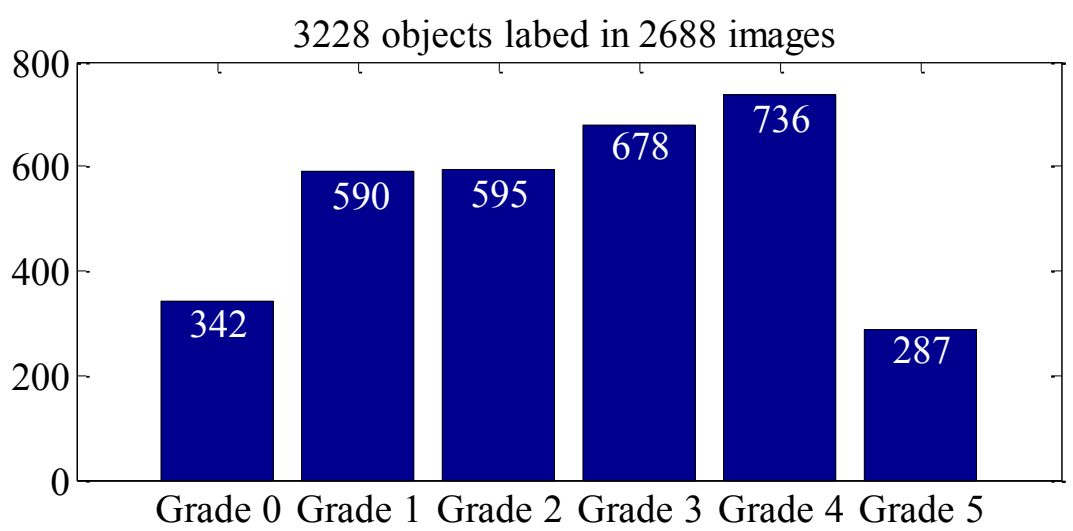

Fig. 1. A total of 3288 ground truths where established for 2688 DF dataset, where some images may contain multiple ulcers.

In the DF dataset, image resolution ranged from 1322x1080 to 3264x2448. In order to reduce computational cost and create efficient training data, we resized all images to 640x640. The Wagner grade classification of each DF image in the datasets was performed strictly in accordance with the patient's clinical diagnosis. A rectangular region of interest (i.e. including the ulcer and its surrounding areas) for the DF dataset was marked by three professional endocrinologists using labelImg(https://githubcom/tzutalin/labelImg), a graphical image annotation tool, and the object bounding boxes were labelled. We chose the entire foot because grade0 in the DF dataset does not differ significantly from healthy feet. The medical experts depicted a total of 3288 ground truths (some pictures with more than one DFU in the DF dataset). As depicted in Fig. 2, the DF image was marked with the corresponding class and bounding box, and an Extensible Markup Language (XML) file was generated. Example images of each Wagner grade are shown in Fig. 3. The Boxes, Classes, and Labelled area information we obtained from parsing the XML files are shown in Fig. 4. In the DF dataset, the number of bounding boxes and classes for each image is only one. The total labelled area of grade0 is the largest, followed by grade5.

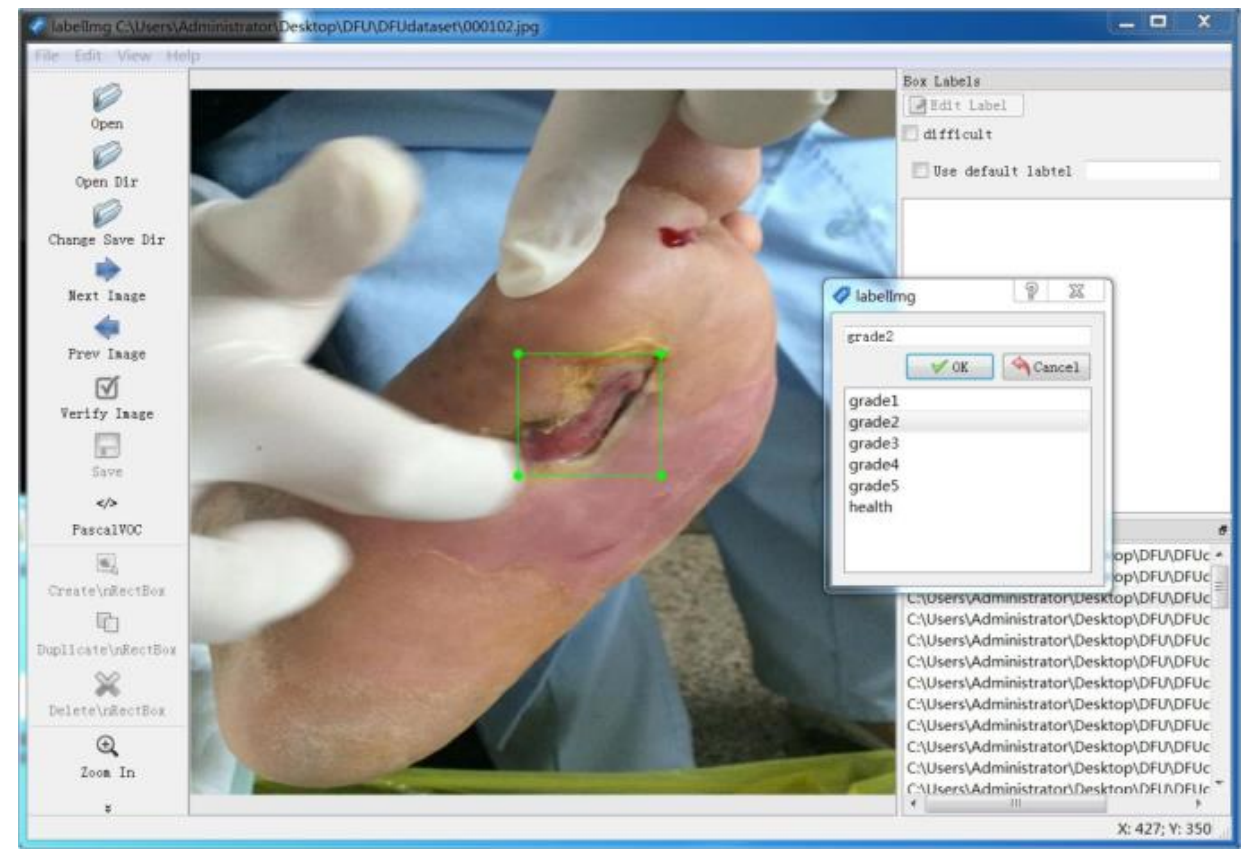

Fig. 2. Example of depicting the ground truth on the DF dataset using labelImg.

### B. Object Detection Algorithm Based on Deep Learning

In recent years, due to the powerful feature fitting capabilities of convolutional neural networks, deep learning-based methods have become widely prevalent in medical imaging, such as DFU [7, 15] and retinopathy [18]. The core tasks of computer vision include object detection, image classification, and semantic segmentation [19]. Compared to image classification, the object detection task includes the identification and location of the target [20], and the performance evaluation metrics are measured by the mean average accuracy. We propose to use object detection algorithms to develop the computer-aided diagnosis system for Wagner Ulcer Classification System.

In the past decade, object detection algorithms have been developed rapidly. In 2012, Krizhevsky et al. utilized Convolution Nural Network (CNN) to win the Image Classification Challenge (ILSVRC2012) [21], which spurred a craze for using CNNs for computer vision tasks. Traditional object detection methods can be divided into two classes. The first type is the method of sliding windows and hand-crafted features by experts. For different categories, variant features need to be designed. For example, HOG (histogram of gradients) and Support Vector Machine [22], the shortcomings of this type of algorithms are obvious that they have weak generalization abilities and the sliding window strategy used in the detection is very time-consuming. The second class is based on the method of the regional proposal. For example, in 2013, Uijlings et al. [23] proposed a selective search (SS) method, which achieves a trade-off between the quality and quantity of the generated window, so that the speed of the algorithm was greatly improved.

Deep learning based on CNNs can reduce the dependence on handcrafts features effectively, mainly because the deep neural network can automatically learn combinations from the underlying features to high-level features. Currently, object detection algorithms based on deep learning can achieve good performance in face, pedestrian, and general detection task. The features obtained through the deep neural network have strong transferability, which allows the utilisation of transfer learning methods.

The state-of-the-art object localization models can be divided into two types of pipelines. One is called multiple-stage models, which need to generate region proposals in advance, and then perform fine-grained object detection. Classical representative models belong to the R-CNN series, such as Faster R-CNN [24], R-FCN [25], Mask-RCNN [26], etc. The second category of models are single stage models, which extract features directly in the network to predict the classification and location of objects. Classical representative models include the Single Shot Detector (SSD) [27], YOLOv1 [28] etc. As shown in Fig. 5, a convolutional layer composed of a standard CNN, such asResNet-101 [29], MobileNetv2 [30], is utilized as a feature extractor to get features from the input pictures as feature maps that are applied to identify objects in the input picture with elaborate attention to the ground truth area. In addition, single stage algorithms predict classes directly and bounding box regression simultaneously, without the need for a region proposal. In multiple-stage algorithms, these feature maps serve as input for the region proposal generator and classification and regression of Region of Interest (RoI). In these cases, the regional proposal is created by a small network sweeping the feature map in a sliding-window strategy to find a specific area including the object. Then, all RoI boxes composed of feature maps generated by the region proposal generator are input to the RoI pooling layer to resize to the same size for the classifier, as the ROI boxes have different sizes. Then, the RoI boxes output from the RoI pooling layer are employed for classification and regression of the bounding box.

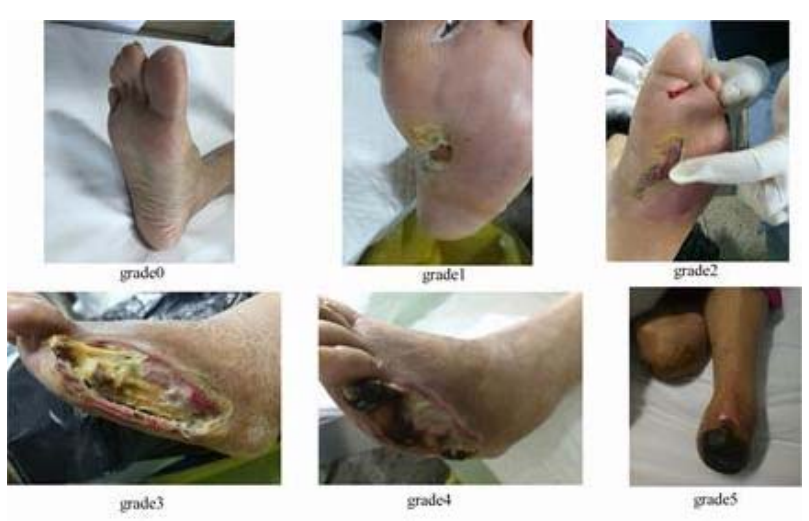

Fig. 3. Illustration of high-resolution Wagner ulcer classification images.

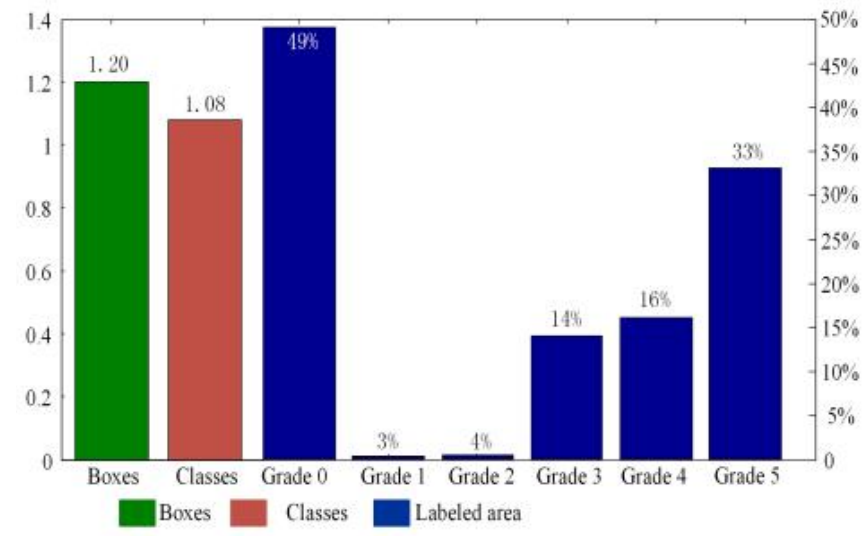

Fig. 4. Ground truth information of DF dataset, where Boxes and Classes respectively represent the average number of bounding boxes and the number of categories of each image in the DF dataset, and the Labeled area indicates the proportion of the average labelled area of each class.

In this paper, Faster R-CNN, SSD, and YOLOv3 are selected as the object localization models as representatives for single and multiple stage-pipelines, respectively. ResNet-101 [29], MobileNetv2 [30], Darknet-53 are selected for the classification algorithms of the above object location models [31]. Faster R-CNN was proposed by Ren et al. in 2015 [24]. It introduced the concept of region proposal networks (RPN), using neural networks to learn the generated region proposal. Since RPN and region-of-interest (RoI) Pooling share the previous convolutional neural network, which greatly reduces the parameter amount and prediction time, it is considered to be the first algorithm that implements end-to-end training. The detection flow chart of Faster R-CNN is shown in Fig. 6.

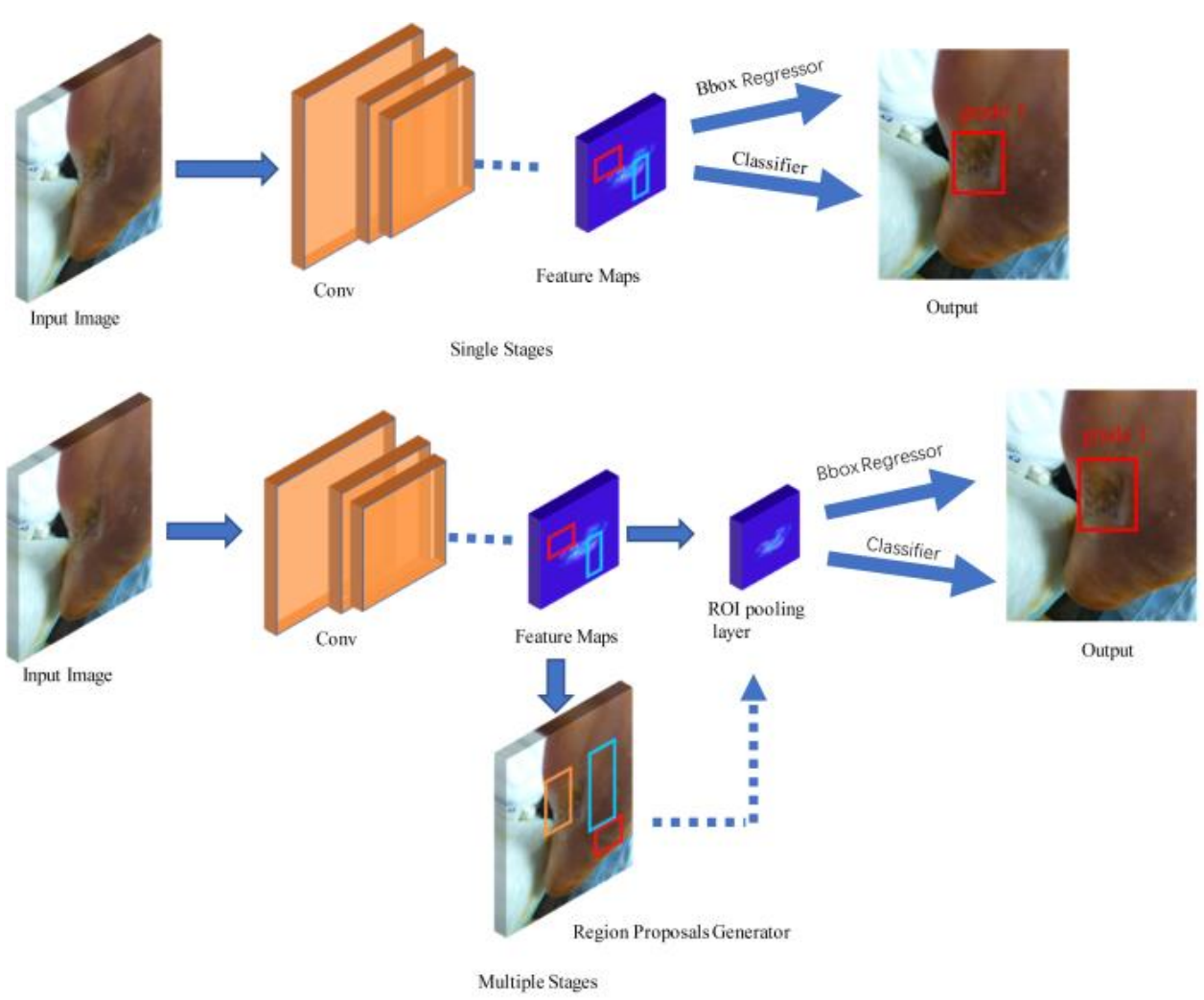

Fig. 5. The difference between single-stage and multiple-stage models is that multiple stage models have a process of region proposal. Conv refers to a convolution layer, and Bbox regressor refers to the Bounding box regressor.

Redmon et al. proposed the YOLOv3 models in 2018 [31], which offers speed and precision improvements over YOLOv1 [28]. Compared with the previous versions, it mainly adjusts the network structure and adds a Residual Block to better acquire object features, depicted in Fig. 7.

Compared with the R-CNN series and the YOLO series object detection algorithms, the Single Shot Detector (SSD) proposed by Liu et al. improves on YOLOv1 [27-28], and its precision is greatly improved while maintaining a relatively fast running speed. SSD completely abandons the process of proposal extraction and feature resampling and utilizes multiple convolutional feature maps for bounding box regression and object label prediction. In order to enable the model to handle different target sizes, it uses different scales and different proportions of default boxes on different size feature maps, and also adds the anchor box adopted by Faster R-CNN [24]. SSD combines the features extracted by Feature maps of different sizes to predict the objects of different sizes [27], as shown in Fig. 8. This improves the detection accuracy of small objects to some extent.

In the study, these models' application scenarios were mainly based on mobile and embedded platforms, which requires efficient response speed. At present, the research on model refinement mainly focuses on the following two directions: 1) Compressing a well-trained complex model to obtain a small model. 2) directly design small models for training. In this paper, we employed the second method, and utilised MobileNetv2 [30], a lightweight network model.

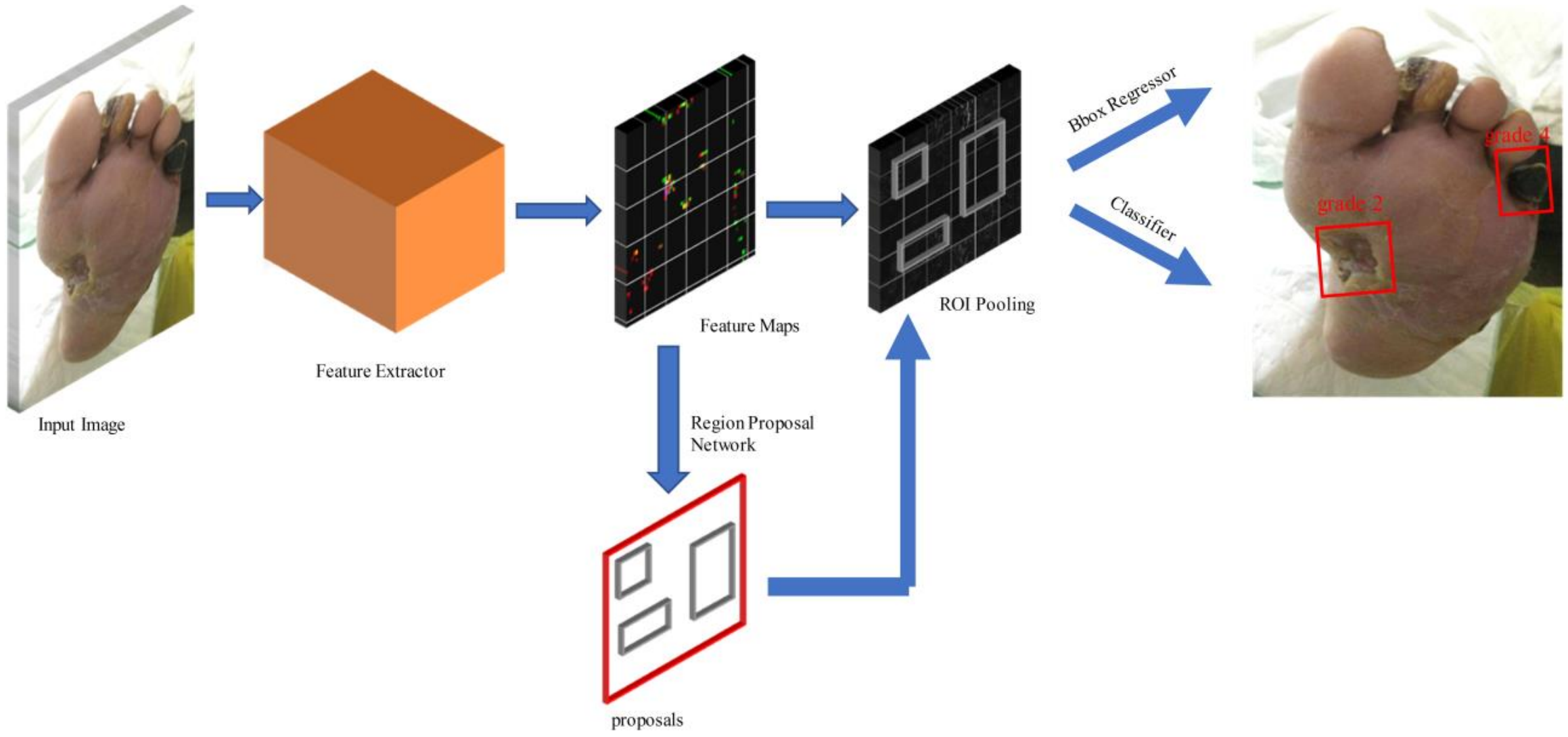

Fig. 6. Detection flow chart of Faster R-CNN.

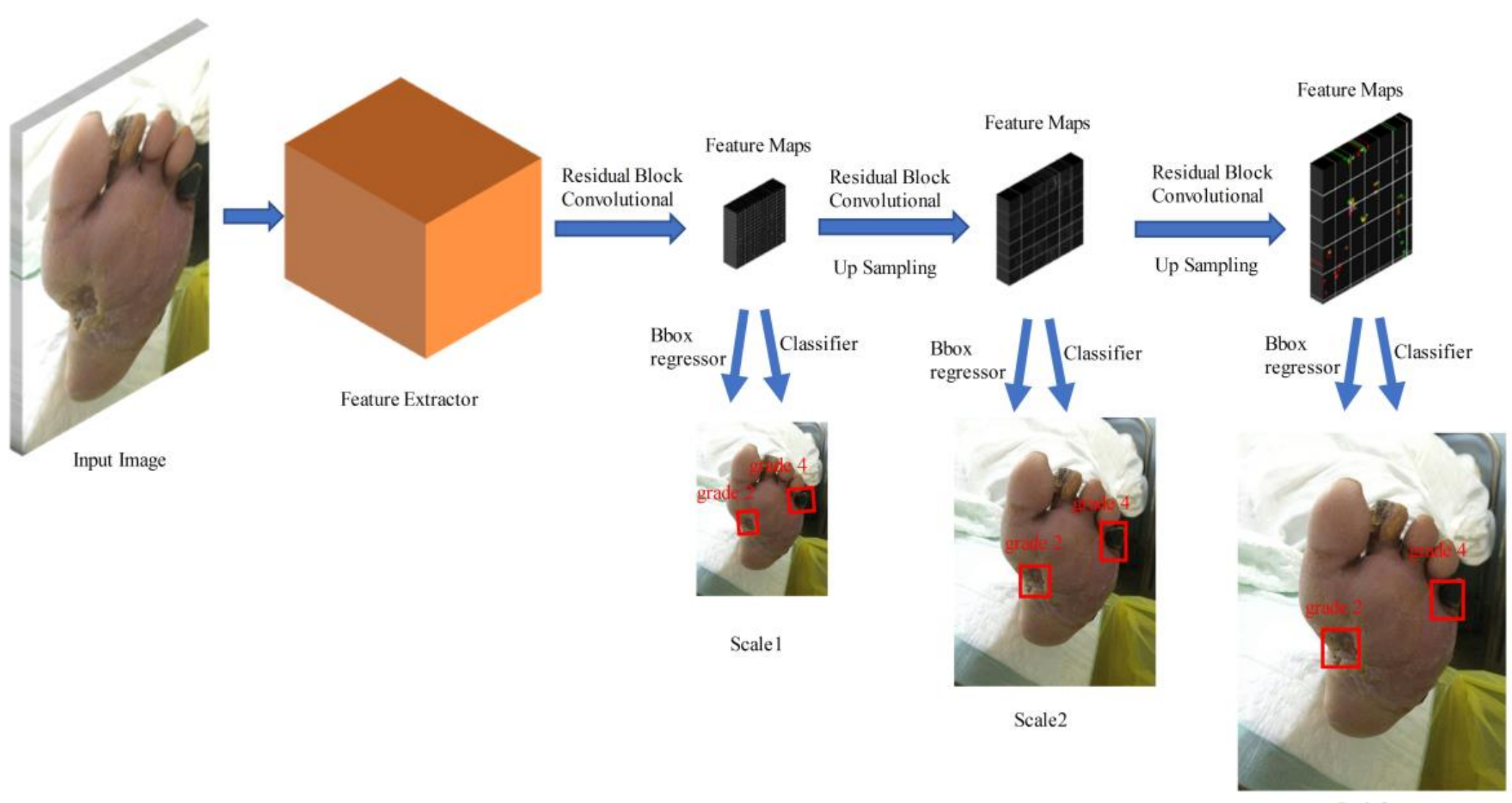

Fig. 7. Detection flow chart of YOLOv3, Compared to Faster R-CNN, the region proposal process is missing. Scale1, Scale2, Scale3 respectively represent the scale of detecting a small, medium, or large object.

The focus of the MobileNetv2 network is to optimize the latency while taking into account the size of the model. Its core idea is to use a depth-wise convolution operation. Under the same number of parameters, the calculation can be reduced by several times compared with the standard convolution operation, so that the computing speed of the network is increased. Compared with the MobileNet network, ResNet-101 i.e. Residual Networks 101, focuses on precision, and the amount of parameters generated after training is very large [29]. Darknet-53 uses a full convolution network that replaces the pooling layer with a convolutional operation of step size 2 [31], and adds a residual unit to avoid gradient disappearance.

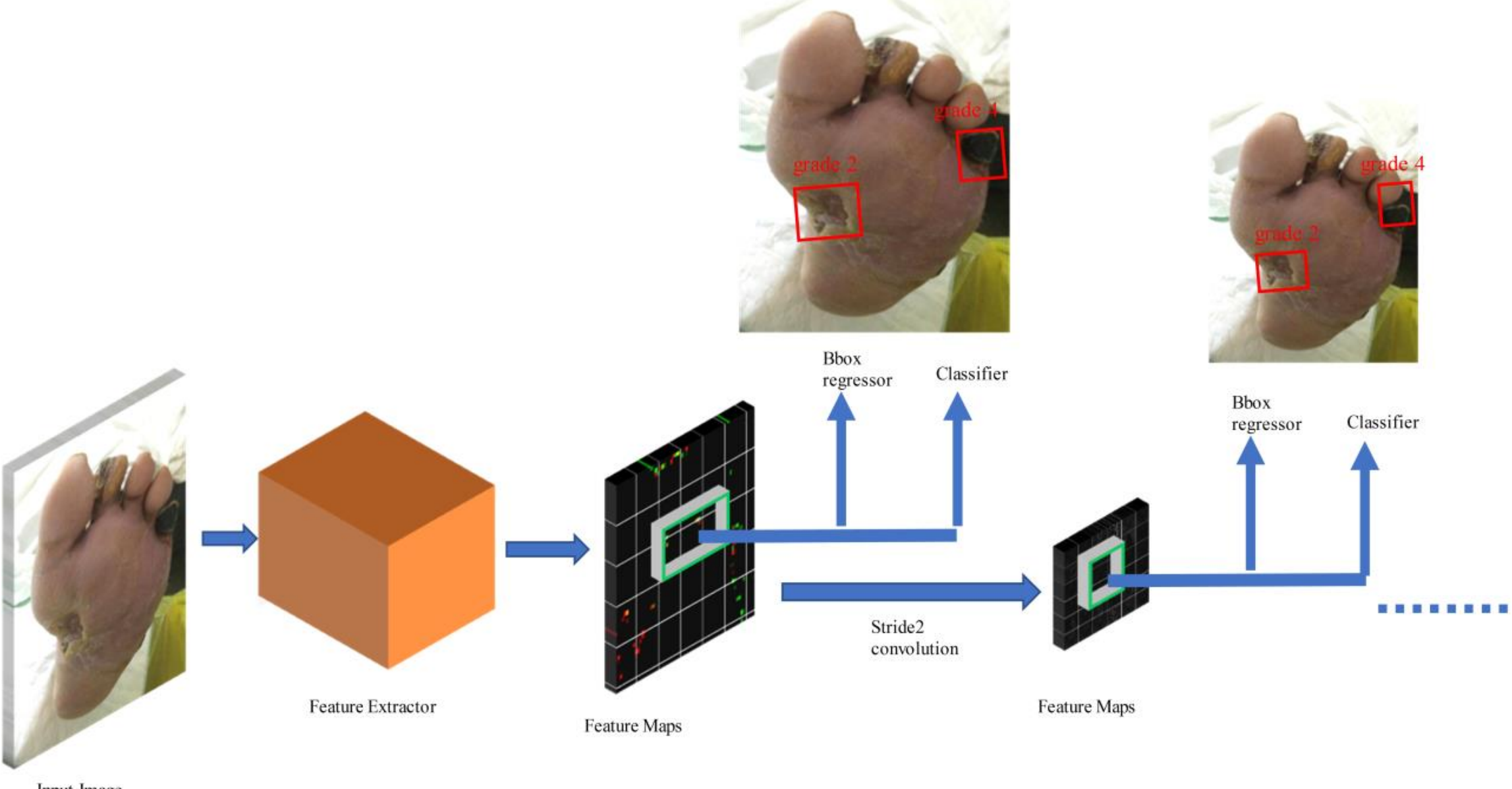

Fig. 8. Detection flow chart of SSD [27]. Each feature map generated by convolution will perform bounding box regression and object class prediction.

### C. Traning Models on DF Dataset

For the DF dataset, we applied stack methods to improve the performance of the originated YOLOv3 model [31], such as Image Mixup [32], Classification Head Label Smoothing [33], Cosine learning rate and commonly used data augmentation methods [34]. In the next section, we will elaborate on the methods used.

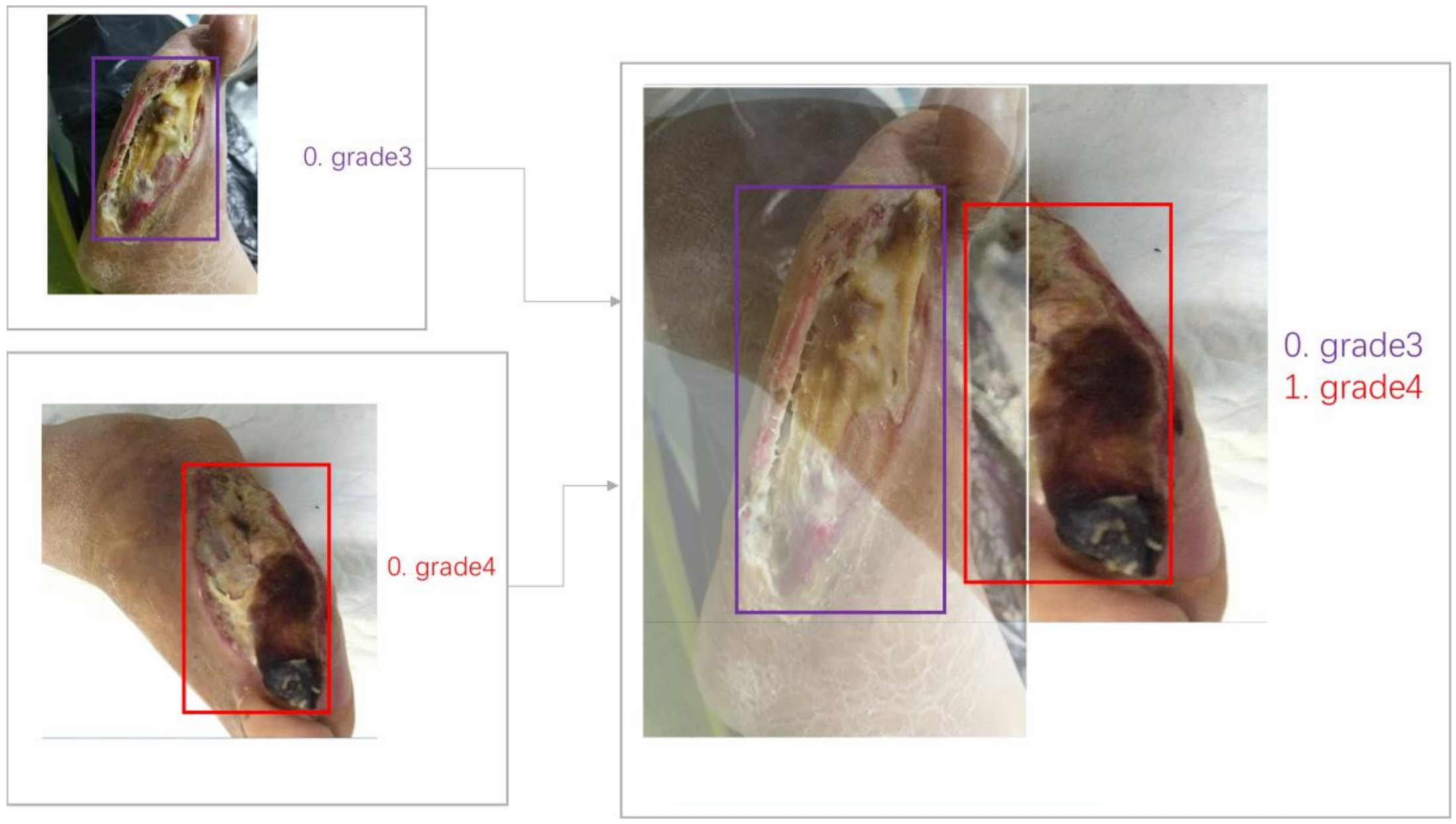

Fig. 9. After two DFU picture pixels are mixed, the ground truth are merged with new samples.

*1) Common data augmentation methods*

In this paper, DF dataset has only 2,688 images, we use data enhancement to complement the DF images to avoid over-fitting during the model's training. A more generalized DF detection model is generated through data augmentation to adapt to the complex real environment. Current mainstream data augmentation methods include geometric distortion (i.e. random cropping, flipping, scaling, etc.) and photometric distortion (i.e. regulating brightness, saturation, etc.).

*2) Image Mixup Method*

The earliest purpose of the visually coherent image mixup method is to solve the problem of disturbance rejection and is very effective in classification networks [34], which is also a way of data enhancement. We were inspired by the visual coherence image mixup experiment of Zhang et al. [35], which was also introduced in our experiments. From the perspective of algorithm, mixup method can be described as follows, we randomly opt two pixels points $(x_i, y_i)$, $(x_j, y_j)$ from two image, respectively, and then mix into a new pixel point with mixup method of Eq. 1 and Eq. 2.

$$\hat{x} = \lambda x_i + (1-\lambda)x_j \quad (1)$$

$$\hat{y} = \lambda y_i + (1-\lambda)y_j \quad (2)$$

where $\lambda \in [0, 1]$, which is randomly generated by Beta (alpha, alpha) distribution. Only the new sample $(\hat{x}, \hat{y})$ is applied for training, i.e. the object label is merged by the new sample. As shown in Fig. 9, the DF picture pixels are mixed in a certain ratio. Here we use the geometry-preserved alignment of mixed images to avoid distortion of the image during the initial steps [35]. We employ Beta(1.5, 1.5) for the DF pictures mixup.

### D. Learning Rate Mode and Label Smoothing

In the training process of the neural networks, the hyperparameter, such as learning rate, is very important. For example, the value of the learning rate impact directly the training time and whether or not to converge the model. One effective solution to the above problem is to define a learning rate that is attenuated with the number of iterations, taking into account training efficiency and stability. The step learning rate schedule is currently a widely used learning rate strategy, which multiplies the predefined epochs or iterations by a number of [0, 1] to achieve the learning rate attenuation. However, it can cause dynamic changes in the learning process every time the learning rate changes [35]. Compared with step learning, cosine learning rate decay coupled with warm restart strategy uses the cosine function as the periodic function and initializes the learning rate at the maximum value of each cycle [34], making the learning rate excessively smoother. In YOLOv3, the learning rate mode was set to the step learning rate schedule and the cosine learning rate to compare the performance of the two learning rate modes on the DF dataset.

Label Smoothing can be seen as a regularization method that blends the ground truth distribution [33]. By adding noise to the output label (y), the model is constrained and the degree of over-fitting of the model is reduced. This processing reduces the model's excessive trust in the label, and can achieve better results for data with less accurate labels [33]. Our experimental results show that the addition of Label Smoothing has a certain improvement in the accuracy of the model in the training of DF dataset.

### E. Model Performance Evaluation Metrics

The model application scenario is on portable devices, which is convenient for clinical real-time diagnosis. Therefore, the model requirements include not only high precision, but also low latency. Combining the above considerations, the experimental results evaluation index is: mean average precision (mAP) defined in the Pascal VOC [36], speed and size of the model. Mean average precision is a common measure of object detection tasks. Speed represents the time required for inference of a single image. The model size is an important performance metric for mobile devices with limited storage space. In order to facilitate the deployment of the trained model to the smartphone, we freeze all the trained model files, i.e. the files in checkpoint format, into the protocol buffer (PB) format. The model size refers to the size after freezing the trained model.

## III. Results and Discussion

For the experiment, the deep learning framework of the three detector meta-structures was based on TensorFlow, which was accelerated using the GPU. The SSD with MobileNetv2 and Faster R-CNN with ResNet-101 models we used are based on the TensorFlow object detection model API [37], an open-source framework that provides rich object detection models that include a variety of backbone networks, such as FPN, ResNet-101, ResNet-50, MobileNetv2, etc. We randomly divided the whole dataset into a training set (70%), a validation set (10%), and a test set (20%). We used a 5-fold cross-validation method to evaluate the DF dataset to ensure that each picture in the entire DF dataset could be evaluated. Specifically, in each fold we included 1882 pictures of the training set, 268 pictures of the validation set, and 538 pictures of the test set. All experiments were carried on a Dell T640 computer equipped with Intel Xeon (R) Silver 4110 (2.2GHz) CPU, 64-GB RAM, Tesla V100 GPU, 32GB-RAM, and Linux centOS7 operating system.

We initialized the network with pre-trained weights employing transfer learning in the coco dataset [38], which consists of 82,783 images in training sets with 90 classes. In the next, we detail the configuration of hyper-parameters, and results of the evaluation on the DF dataset for the three models.

For SSD with MobileNetv2 and Faster R-CNN with ResNet-101 models, we limited the training steps to 200,000, which we experientially found to be adequate enough to train the DF dataset. In the training process, we attempted a variety of learning rates to ensure that the model training results are optimal. For other hyper-parameters and configurations, we followed the default settings for both models [37]. For SSD, we set both height and width for the fixed shape resizer as 300, the score threshold as 1e-8, iou threshold as 0.6. We used random horizontal flip and random crop as data augmentation options.

For Faster R-CNN, the iou threshold and score threshold settings were consistent with SSD, but we only used random horizontal flip for data augmentation.

For YOLOv3, we set the training process to 200 epochs and used k-means clustering to determine the bounding box priors. More specifically, the nine bounding box priors were (12x10), (16x28), (30x26), (32x43), (51x68), (79x126), (141x86), (233x192), (284x346). The first three bounding box sizes are for small objects, the middle three bounding box sizes are for medium objects, and the last three bounding box sizes are for big objects. During training, the step learning rate schedule for YOLOv3 reduced the learning rate by a ratio of 0.1 at 160 and 180 epochs [34]. We used random shape training instead of fixed dimensions, i.e. in each epoch, models were randomly fed from the predefined resolutions W, where W = {320x320, 352x352, 384x384, 416x416, 448x448, 480x480, 512x512, 544x544, 576x576, 608x608}. We also performed random translates, random crops and random horizontal flips for data augmentation.

Next, we report the performance measures of stacking tricks on the DF dataset. The details are listed in Table I. By stacking all these refinements on YOLOv3 models, we can achieve a performance gain of up to 1.36%. It is worth mentioning that although the mixup experiment achieves an absolute accuracy improvement of 0.87%, some categories like the averages' accuracy are degraded, such as for grade1, grade2. We suspect that this is caused by the high inter-class similarity between Wagner grades. This blending of high inter-class similarity of DF image pixels can lead to misjudgment or missed judgment.

Then, we report the performance evaluation of three models for DF dataset on 5-fold cross validation. The details presented in Table II. Overall, all the models report promising performance, but the Faster R-CNN model does not achieve the expected results, as its mAP is only 0.95% higher than the SSD model. According to the Tensorflow detection model zoo(https://github.com/tensorflow/models/blob/master/research/object_detection), in the coco dataset [38], Faster R-CNN with the ResNet-101 model is 10% higher than SSD with MobileNetv2 model in the mAP performance index, so a larger performance difference would have been expected. The application of the low-complexity MobileNetv2 for classification experiments was mainly to reduce computational complexity. In contrast, high-complexity networks are prone to overfitting and inferior generalization performance for small-size datasets, such as ResNet-101. In terms of speed and model size, SSD with MobileNetv2 ranked first, refinements on YOLOv3 models achieved almost equal performance being only 4 seconds slower than SSD with MobileNetv2, while Faster R-CNN with ResNet-101 ranked last. In mAP performance measures, refinements on YOLOv3 models outperformed the other two models by a distinctive margin, reaching 91.95% of mAP, which is 2.54% higher than SSD models and 1.59% higher than Faster R-CNN models. Thus, on DF dataset, refinements on YOLOv3 models achieved a good trade-off of speed/precision.

A few instances of the test by three models trained are shown by the Fig. 10. Overall, the three models localization accurate with high confidence, but we found that some images still have missed detections. For example, in the fourth row, the SSD model failed to detect a grade 4 ulcer and a grade 1 ulcer. It is worth mentioning that from Fig. 10, we see that the refinements on YOLOV3 models predict that the confidence value of a certain class is significantly smaller than the other two models, which is caused by the addition of the mixup.

The training and testing of the above models were run on servers with GPU-accelerated computing, but for clinical diagnostics, portable devices that can be carried around are more practical.Android studio combined with the TensorFlow deep learning mobile library makes it easy to deploy trained models to portable devices such as Android smartphones. Considering the accuracy and real-time requirements of portable applications, we packaged the refinements on the YOLOv3 models into a prototype android application and deployed them on Android smartphones for real-time DF detection and Wagner grades assessment.

TABLE I. TRAINING REFINEMENTS ON YOLOV3, EVALUATED AT 544X544 FOR DF DATASET ON 5-FOLD CROSS VALIDATION, WHERE DELTA REPRESENTS THE INCREMENT RELATIVE TO THE BASELINE

| **Tricks** | **mAP (%)** | **Delta (%)** | **grade0 (%)** | **grade1 (%)** | **grade2 (%)** | **grade3 (%)** | **grade4 (%)** | **grade5 (%)** |
|---|---|---|---|---|---|---|---|---|
| baseline | 90.59 | 0 | 95.93 | 88.57 | 89.71 | 90.62 | 85.91 | 92.81 |
| +cosine learning | 90.87 | 0.28 | 95.25 | 89.33 | 89.87 | 91.43 | 85.76 | 93.57 |
| +label smoothing | 91.08 | 0.49 | 96.52 | 90.15 | 90.94 | 89.13 | 86.27 | 93.48 |
| +mixup | 91.95 | 1.36 | 96.88 | 88.76 | 89.79 | 91.26 | 89.58 | 95.41 |

TABLE II. PERFORMANCE EVALUATION OF SSD, FASTER R-CNN AND REFINEMENTS ON YOLOV3 ON DF DATASET IN 5-FOLD CROSS-VALIDATION.

| **Model** | **mAP (%)** | **grade0 (%)** | **grade1 (%)** | **grade2 (%)** | **grade3 (%)** | **grade4 (%)** | **grade5 (%)** | **Size of Model (MB)** | **Speed (ms)** |
|---|---|---|---|---|---|---|---|---|---|
| SSD-Mobilev2 | 89.41 | 93.53 | 87.24 | 88.58 | 87.95 | 86.78 | 92.39 | 18.6 | 27 |
| Faster R-CNN-Res101 | 90.36 | 96.33 | 86.79 | 88.43 | 87.41 | 89.38 | 93.80 | 181.8 | 104 |
| refinements on YOLOv3 | 91.95 | 96.88 | 88.76 | 89.79 | 91.26 | 89.58 | 95.41 | 34.8 | 31 |

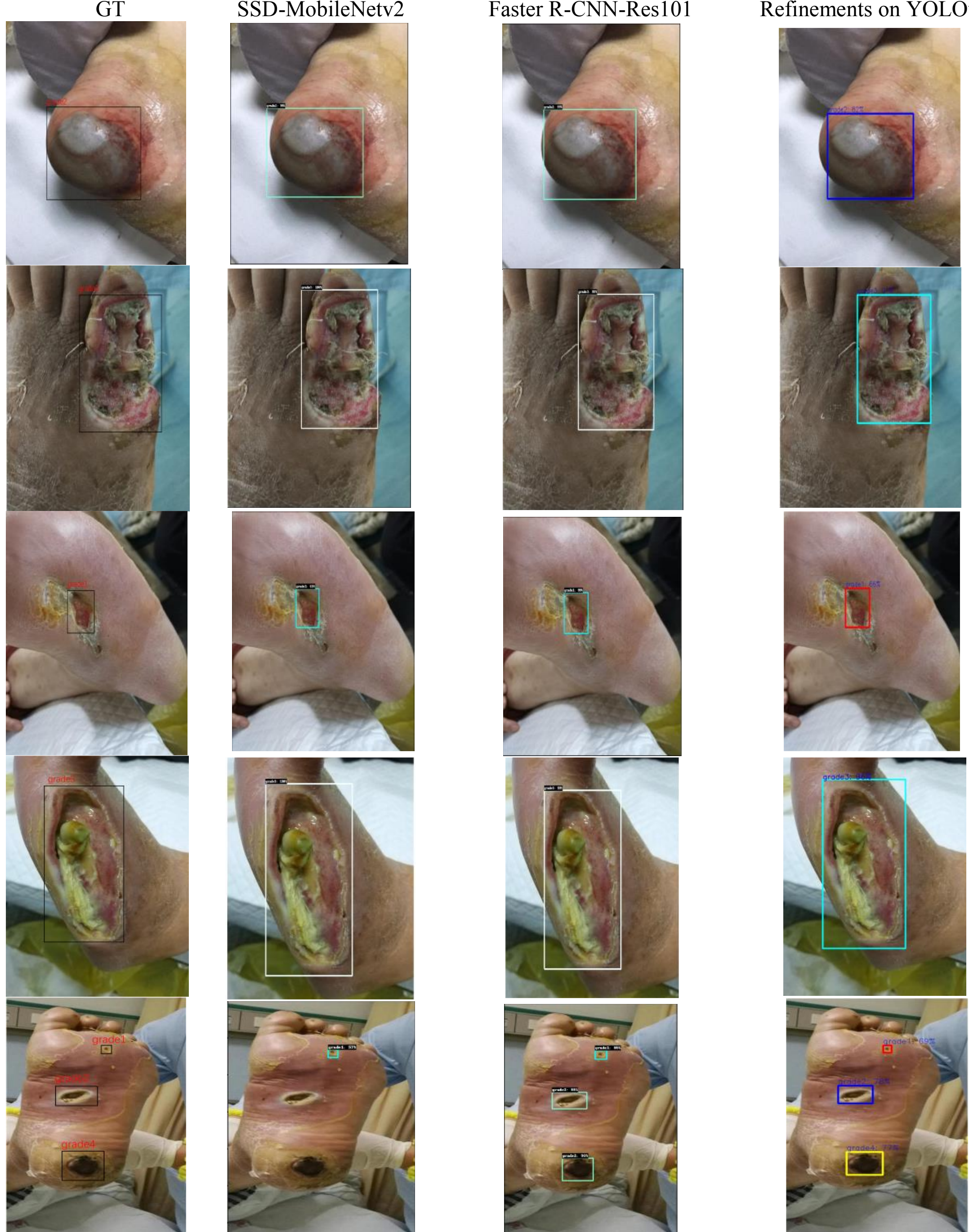

Fig. 10. Results of DF detection using three trained models for visual performance comparison. GT represents the ground truth.

To validate the real-time performance and accuracy of the model, we tested a prototype android application with HUAWEI HONOR10 at the 1st. of FMU. We tested a total of 37 DF patients, including 10 grade0, 6 grade1, 6 grade2, 11 grade3, 2 grade4, and 2 grade5. The test achieved promising localization performance and no grade misjudgments were made by the application. Some examples are shown in Fig. 11. where the second column represents the snapshot of real-time detection and localization by the android application. It is worth mentioning that our model is only for patients who have been already diagnosed with a DF. Using the three trained models described above, we examined 20 pictures of healthy feet and found that the model determined all healthy feet as grade 0, mainly because grade 0 had no significant difference from healthy feet in appearance.

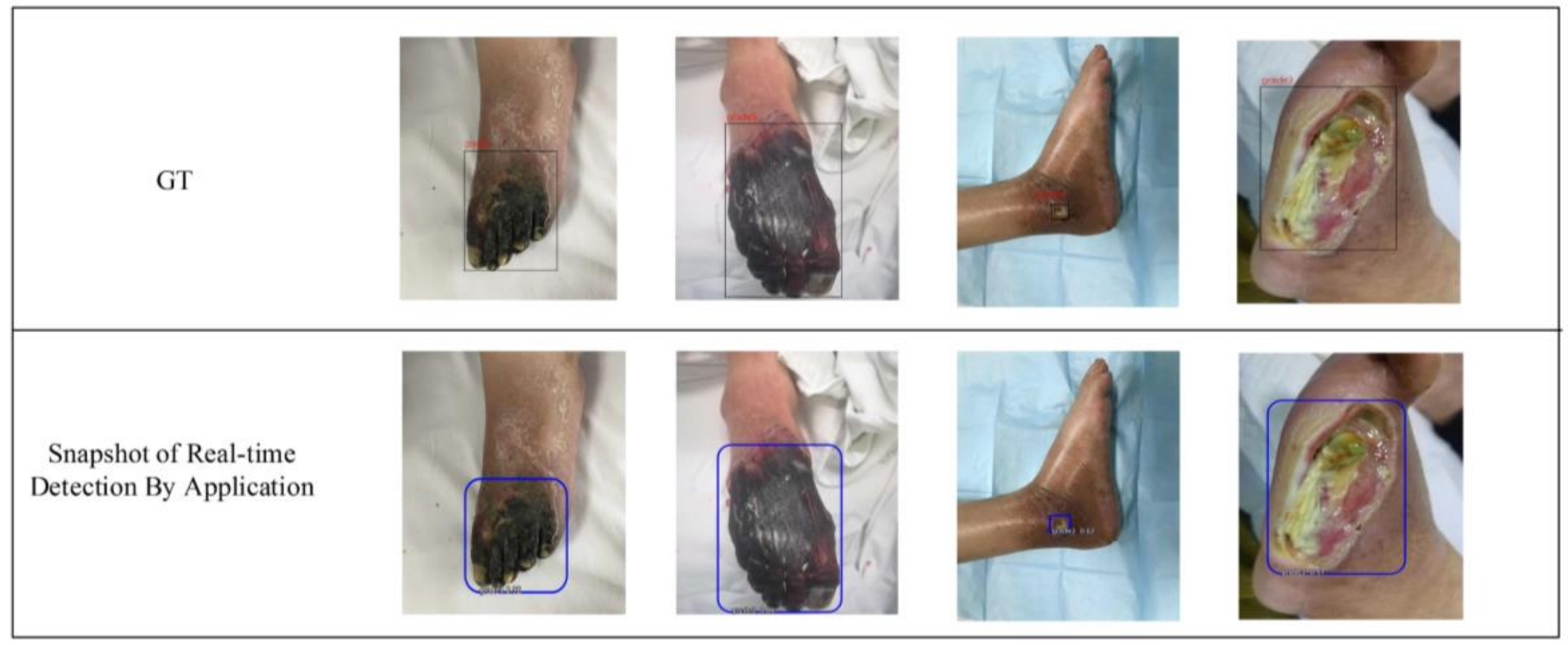

Fig. 11. Examples of clinical real-time detection of Wagner grades using the refinements on YOLOv3 models.

## IV. CONCLUSION

In this work, we used three different object detection algorithm to realize the classification and analysis of Wagner ulcer for the first time. To construct a robust algorithm model, we asked three experienced medical experts to delineate the DF and its surrounding area on 2688 images in the form of a bounding box. We then used a bag of tricks on YOLOv3. As shown in Tables I and II, by using refinements on YOLOv3, we have achieved an optimal speed/precision trade-off than SSD and Faster R-CNN models. From Fig. 10, we also found that because of the mixup method, the refinements on YOLOv3 predict the confidence value of a certain class is obviously smaller. Finally, we packaged the refinements on YOLOv3 model into an Android application and deployed it in a HUAWEI HONOR10 (Android Phone) for real-time Wagner grades detection. As can be seen from Fig. 10, the prototype application achieved good performance. In summary, we developed real-time mobile detection and localization Wagner grades of systems that can provide an effective evaluation for DF tissue analysis and healing status, which may shift future clinical treatment method of DF. Although the algorithm has achieved good performance, the dataset in this paper comes from Asians, so the algorithm may have low mAP in identifying caucasian or blacks. In the future, the work of this paper can further expand the caucasian or blacks dataset, thereby further improving the mAP and robustness of the algorithm.

## ACKNOWLEDGEMENT

The authors acknowledge the Fujian Province STS Projects (Grant: 2019T3008 and 2019T3009), the Diabetes Fund from Chinese Society of Microcirculation (Grant:TW-2018P002) and the Central Government Special Funds for Local Science and Technology Development (Grant: 2018L3007). Moreover, the contributions from Dr. Manu Goyal are greatly appreciated.